\documentclass{article}

\usepackage[preprint,nonatbib]{neurips_2023}

\usepackage[utf8]{inputenc}
\usepackage[T1]{fontenc}   
\usepackage{hyperref}      
\usepackage{url}           
\usepackage{booktabs}      
\usepackage{amsfonts}      
\usepackage{nicefrac}      
\usepackage{microtype}     
\usepackage{xcolor}        
\usepackage{tikz-cd}
\usepackage{enumitem}
\usepackage{fbox}

\usepackage{amsthm}
\theoremstyle{plain}
\newtheorem{theorem}{Theorem}[section]

\theoremstyle{definition}
\newtheorem{definition}[theorem]{Definition}

\theoremstyle{remark}

\title{CatCode: A Comprehensive Evaluation Framework for LLMs On the Mixture of Code and Text}

\author{
  Zhenru Lin\\
  IIIS, Tsinghua University\\
  \texttt{lzr22@mails.tsinghua.edu.cn}
  \And Yiqun Yao\\
  Beijing Academy of Artificial Intelligence\\
  \texttt{yqyao@baai.ac.cn}
  \And Yang Yuan\thanks{Corresponding author.}\\
  IIIS, Tsinghua University\\
  Shanghai Artificial Intelligence Laboratory \\
  Shanghai Qi Zhi Institute\\
  \texttt{yuanyang@tsinghua.edu.cn}
  }

\begin{document}

\maketitle

\begin{abstract}

Large language models (LLMs) such as ChatGPT are increasingly proficient in understanding and generating a mixture of code and text. Evaluation based on such \emph{mixture} can lead to a more comprehensive understanding of the models' abilities in solving coding problems. However, in this context, current evaluation methods are either limited in task coverage or lack standardization. To address this issue, we propose using category theory as a framework for evaluation. Specifically, morphisms within a code category can represent code debugging and transformation, functors between two categories represent code translation, and functors between a code category and a natural language category represent code generation, explanation, and reproduction. We present an automatic evaluation framework called \textbf{CatCode} (\textbf{Cat}egory \textbf{Code}) that can comprehensively assess the coding abilities of LLMs, including ChatGPT, Text-Davinci, and CodeGeeX.

\end{abstract}

\section{Introduction}

The success of large language models (LLMs) as programming assistants has been widely acknowledged, with their higher proficiency demonstrated in various coding tasks such as code generation \cite{alphacode2022,intelliCode2020Svyatkovskiy}, code explanation \cite{gpt3explain-MacNeilTMBRH22}, and code translation \cite{zhu2022multilingual}, among others. 
For instance, AlphaCode ranked in the top 54.3\% in simulated evaluations on recent programming competitions on the Codeforces \cite{alphacode2022}. The Codex-powered programming tool Copilot \cite{codex} serves as an effective tool for generating and editing code.
Undoubtedly, LLMs exhibit remarkable capabilities when addressing different coding scenarios individually. 
However, what truly sets the recent LLMs apart is their ability, akin to ChatGPT-like models, to comprehend and align with human intents by processing a \emph{mixture} of natural language and code. This unique attribute significantly lowers the entry barrier for users, leading to their widespread adoption and notable achievements.

To assess the coding ability of LLMs, numerous efforts have been made by researchers. However, current evaluation methods are either limited in task coverage or lack standardization.
Match-based automatic frameworks such as CodeBLEU \cite{codebleu2020} rely primarily on similarity scores to evaluate the quality of code. However, such frameworks often fail to capture the nuances of code functionality and meaning. 
Execution-based evaluation methods, for example, MBXP \cite{mbxp} and MultiPL-E \cite{cassano_multipl-e_2023}, can evaluate the function accuracy of code, but they primarily focus on code generation and translation tasks.
Task-based evaluation frameworks like CodeXGLUE \cite{lu2021codexglue} offer a comprehensive approach but lack standardization due to variations in datasets, task formulations, and APIs.
Consequently, we still lack an evaluation framework to adapt to the context of a mixture of natural language and code, and there is a need to establish a comprehensive evaluation framework that not only supports diverse and novel task definitions, but also provides a standardized approach for evaluating the model under such a mixture context.

We aim to establish a comprehensive theoretical framework that can be open-sourced and applied to essential coding tasks, providing standardized automatic evaluation metrics.
Designing an extensive and standardized evaluation framework becomes a challenging task in the absence of theoretical guidance.
To address this need, we seek a theory that can effectively express the structural aspects of code, language, and their interrelationships. 
In this context, we find category theory, a branch of mathematics, to be particularly suitable for describing relationships between various elements. 
Furthermore, there are existing applications for employing category theory to describe code and language, making it an appropriate choice as a framework. 

By utilizing category theory's mathematical abstractions, we can gain insights into the relationships among different programming languages (PLs) and natural languages (NLs). 
We consider PLs and NLs as categories, functionally equivalent programs as objects, and leverage functors and morphisms to capture the object relations within and across these categories. 
This gives a unified framework for describing functional equivalence, which not only works within a single PL, but also among different PLs, and even between PLs and NLs. 
By learning the morphisms within a programming language category, the model can grasp the similarities and differences between code snippets. 
Additionally, by acquiring knowledge of the functors between categories, the model can understand the relationship between different programming languages and natural languages.

\begin{figure}[t]
    \centering
    \includegraphics[width=\linewidth]{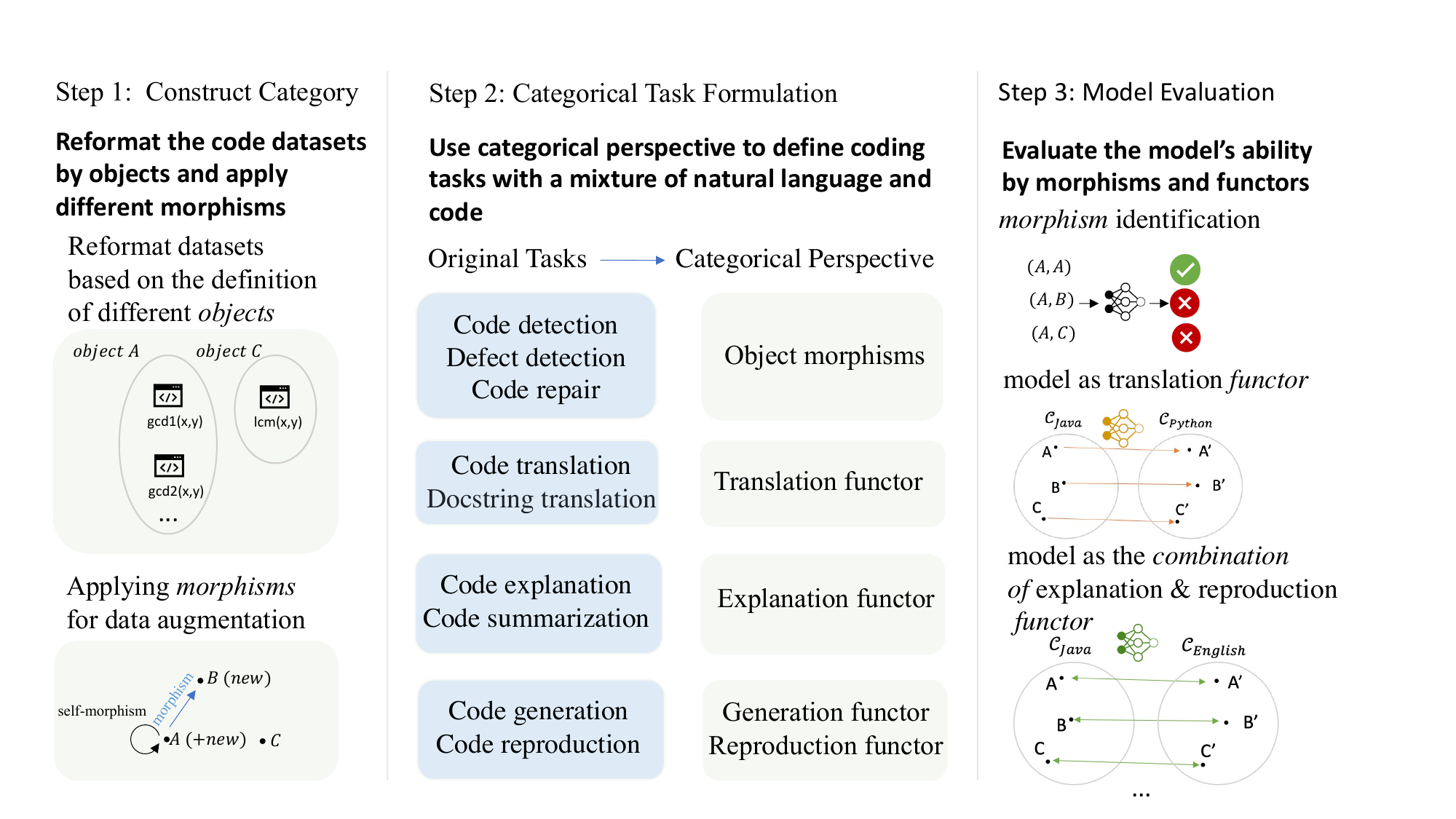}
    \caption{The overall evaluation framework. We use category perspectives to reorganize and transform data, formulate different coding tasks, and conduct model evaluations. }
    \label{fig:evaluation_framework}
\end{figure}

Based on the categorical perspective above, we build a standardized evaluation framework (see Figure~\ref{fig:evaluation_framework}). It is standardized in terms of data definition, task formulation, and APIs. 
This framework can be extended to many code-related tasks as long as a categorical definition is given. In our experiments, we give some examples of common code-related tasks, and assess models such as ChatGPT and Text-Davinci for their capabilities in identifying functional equivalence in code, performing code translation, generating code explanations, and reconstructing code based on explanations. We have observed that these models still struggle to differentiate between the concepts of "functional equivalence" and "similarity" in code. While they demonstrated relatively satisfactory performance in code translation, maintaining functional equivalence between code and its corresponding explanations remains a persistent challenge.

Our main contributions can be listed as follows: 
\begin{itemize}[leftmargin=*]
    \item We introduce \textbf{CatCode}, a novel evaluation perspective for code-related tasks based on category theory, which provides a \emph{comprehensive} framework that encompasses a wide range of code-related task formulations. 
    \item We present a \emph{standardized} automatic evaluation platform within the CatCode framework, that offers a quantitative assessment of the coding abilities of Language Models (LLMs) and can adapt to various datasets and models, which will be publicly available.
    \item We evaluate competitive LLMs, including ChatGPT and Text-Davinci\cite{instructgpt}, providing insights into their strengths and limitations in understanding the \emph{mixture} of NL and code.
\end{itemize}

\section{Methods}

Generally, it is difficult to achieve both comprehensiveness and standardization. 
In Section~\ref{sec:comprehensive}, we show that the categorical perspective, with its emphasis on generalization and abstraction, offers a valuable approach to achieving comprehensiveness in dealing with the mixture of code and natural language.  
In Section~\ref{sec:standardized}, we discuss the significance of standardization and outline strategies for achieving it.

\begin{figure}[t]
    \centering
    \includegraphics[width=0.7 \linewidth]{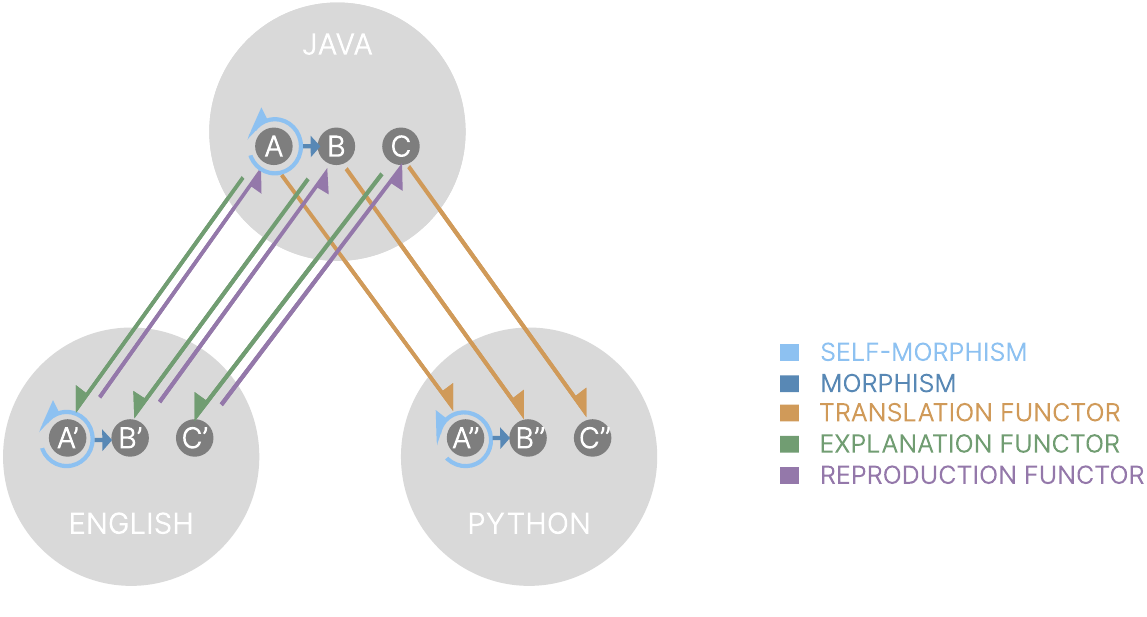}
    \caption{Categorical framework for a mixture of code and NL. $A$, $B$ and $C$ represent different objects, $A$ and $A''$ represent the equivalent object of $A$ in other categories.}
    \label{fig:catcode}
\end{figure}

\subsection{Comprehensive Categorical Perspective}
\label{sec:cat_applied}
\label{sec:comprehensive}
Category Theory has been applied to different fields, ranging from functional programming to logic, software design, and linguistics\cite{cat_science,functionalprogramming,fong2018seven,bradley2018applied}.
Here we provide the basic definitions used in our work and explain their applications to describe code and natural language. Figure \ref{fig:catcode} illustrates those concepts in a \emph{comprehensive} way.

\begin{definition}
\label{def1}
(Category, Object, Morphism). A category $\mathcal{C}$ consists of a set of objects $\mathrm{Ob}(\mathcal{C})$ and a set of morphisms $\mathrm{Hom}_\mathcal{C}(a,b)$, from $a$ to $b$ for every $a,b\in \mathrm{Ob}(\mathcal{C})$. 
Given two morphisms $f \in \mathrm{Hom}_{\mathcal{C}}(a, b), g \in \mathrm{Hom}_{\mathcal{C}}(b, c)$, we define their composition as $g \circ f \in \mathrm{Hom}_{\mathcal{C}}(a, c)$. The composition of morphisms is governed by two axioms:

- \emph{Associativity}: If $f: a \rightarrow b, g: b \rightarrow c$, and $h: c \rightarrow d$ then
$$
h \circ(g \circ f)=(h \circ g) \circ f .
$$

- \emph{Identity}: For every object $x$, there exists a morphism $\mathrm{id}_x: x \rightarrow x$ called the identity morphism for $x$, such that
   for every morphism $f: a \rightarrow b$, we have
$$
\mathrm{id}_b \circ f=f=f \circ \mathrm{id}_a .
$$
\end{definition}

\textbf{Application 2.1.} (Category, Object, Morphism of Code and NL).
For describing the mixture of code and natural languages, we first regard them as objects from different categories. We consider the code categories which contain all executable code in a certain language: $\mathcal{C}_{Java}$, $\mathcal{C}_{Python}$, ... 
We define the natural language categories which contain all description/explanation/summarization of code etc.: $\mathcal{C}_{English}$, $\mathcal{C}_{Chinese}$, ...
For simplicity, let’s call them $\mathcal{C}_1$, $\mathcal{C}_2$, ...

In a category, there are infinitely many different objects. We name these objects $o_1$, $o_2$, ...
Each object contains infinitely many programs, with \emph{the same running outcome for each valid input}. 
We use this definition because we focus on the functional equivalence of code function.
For example, one may have slightly different implementations of quick sorts, but they represent exactly the same function. \footnote{We can also extend the definition of the object to be the set of code with not only the same outcome, but also with the same time or space complexity, but due to the difficulties of automatically evaluating the complexity, this can be extended for future work.}

Based on the definition of objects, we define the morphism between two objects, as ``necessary edits to convert a function to another''. 
In particular, we define the self-morphism for each object, as ``edits that do not change the functionality of the program''. 

\begin{definition}
 (Functor). 
A functor $F$ from a category $\mathcal{C}$ to a category $D$, written as $F: \mathcal{C} \rightarrow D$, maps an object $x \in \mathrm{Ob}(\mathcal{C})$ to $F(x)\in \mathrm{Ob}(D)$; as well as a morphism $f: x \rightarrow y$ in $\mathrm{Hom}(\mathcal{C})$ to $F(f): F(x) \rightarrow F(y)$ in $D$, such that the following two properties hold:
  
- For every object $x$ in $\mathcal{C}, F(\mathrm{id}_x)=\mathrm{id}_{F(x)}$;

- For all morphisms $f: x \rightarrow y$ and $g: y \rightarrow z, F(g \circ f)=F(g) \circ F(f)$.
\end{definition}

\textbf{Application 2.2.} (Functor of Mixture of Code and NL).
We define the functor from $\mathcal{C}_1$ to $\mathcal{C}_2$, to be the transform from one language to another language, but with the same functionality. When it applies to two code categories, it usually represents code translation, and we define it as a ``translation functor''. When the functor is between a code category and a natural language category, it may have many possible meanings, for instance, we can define ``generation function'' from NL to PL that generate code solution to a problem description, ``explanation functor'' from PL to NL that explains a piece of code, and ``reproduction functor'' from NL to PL that generates code snippets based on code descriptions. 

\subsection{Standardized Evaluation Platform}
\label{sec:standardized}

\begin{figure}[t]
    \centering
    \includegraphics[width=0.99\linewidth]{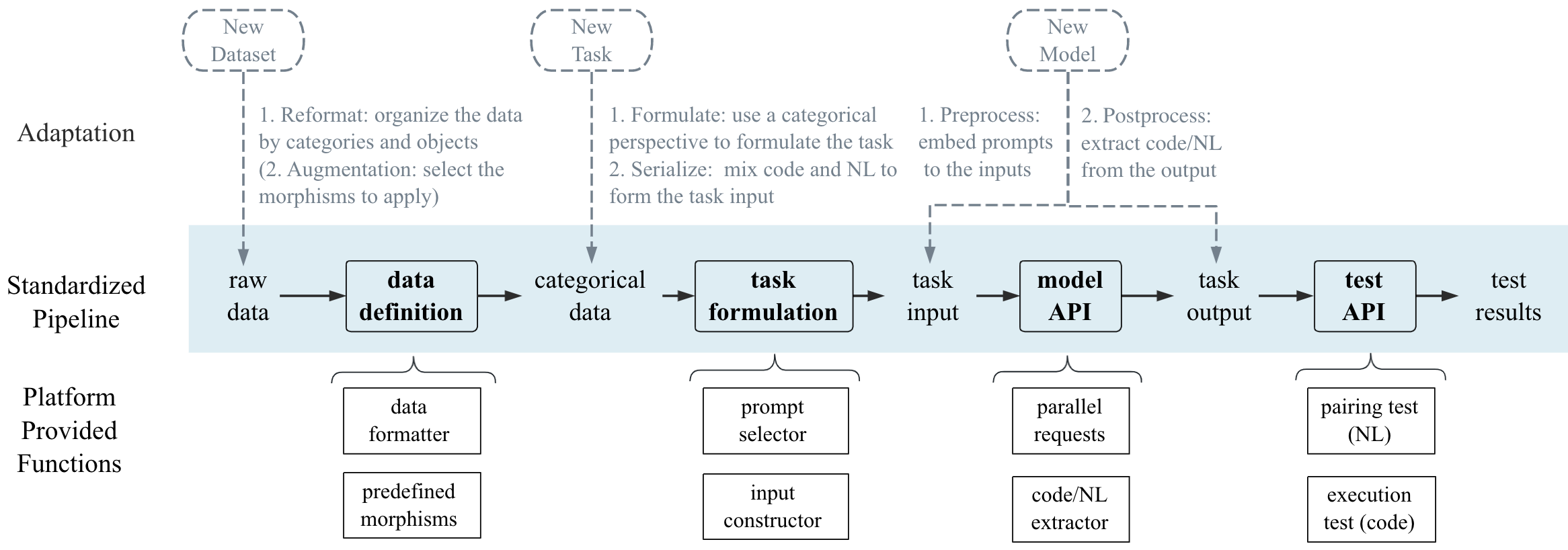}
    \caption{Standardized evaluation platform. The central pipeline offers a consistent approach for all evaluations. Behind the pipeline, we provide a variety of functions to automatically conduct the most important steps. With our platform released, the pipeline can easily accommodate novel datasets, tasks, and models by following the instructions outlined alongside the grey lines.}
    \label{fig:platform_pipeline}
\end{figure}

Figure \ref{fig:platform_pipeline} shows the streamlined process of our standardized evaluation platform. Standardization within the framework is achieved through a focus on three critical aspects: data definition, task formulation, and APIs.

\textbf{Data definition}

The original datasets may consist of a mixture of code and natural languages, so a clear data definition is crucial. The code can be in different languages (Java, Python, JavaScript, etc.), and on different levels (single-line, function-level, file-level). Natural language can be in different relationships with the code (problem descriptions, code comments, demands, etc.). 

By establishing a standardized data definition, we ensure compatibility and facilitate the comparison and integration of different datasets. 

For implementation, we use a ``data formatter'' to rearrange code based on the definition of objects and categories. Additionally, we provide ``predefined morphisms'' and implement them based on \texttt{JavaTransformer}\cite{javatransformer} to automatically apply morphisms to code objects, which makes it easy for data augmentation. 

\textbf{Task formulation}

We use a categorical perspective to formulate diverse code-related tasks using the unified language of objects, morphisms, and functors.

A good task formulation allows for a more generalized and flexible approach to defining more complex code problems, enabling the inclusion of a wide variety of code-related tasks and comparing their similarities and differences. 

Based on the task formulation, we carefully decide what parts of the code and NL should be the model input, ensuring a cleaner setting. Then we use ``prompt selector'' to select a suitable prompt for task description, and use ``input constructor'' to combine the data and prompt as task input.

\textbf{APIs}

After the data is ready, we focus on the standardization of model APIs and test APIs.

Standardized APIs promote transparency, fairness, and efficiency in the evaluation.

By defining a clear and consistent set of APIs, we have integrated OpenAI models into the evaluation process and can test them using parallel requests. For postprocessing the model's output, we enable the extraction and filtering of plain text from the code-text mixture. 
For test APIs, we provide a ``pairing test'' API for evaluating the model's answer based on natural language, and an ``execution test'' API, which connects to \texttt{Mxeval}\cite{mbxp} for compiling and running the tests for a given code.

\section{Experiments}

\subsection{Research Questions and Basic Settings}

In this section, we exhibit some experimental examples of how to use our platform. Meanwhile, we explore the following three research questions (RQs) that correspond to those illustrated in Step 3 of Figure \ref{fig:evaluation_framework}. 

\fparbox{\textbf{RQ1: Can the model exactly capture code functions and identify similarities/differences?}}

(Related tasks: code detection, defect detection, code repair)
 
$\rightarrow$ Categorical perspective: Can the model identify the self-morphisms and other morphisms within the same code category? 

\fparbox{\textbf{RQ2: Can the model translate code between different programming languages? 
 }}
 
 (Related tasks: code translation)

$\rightarrow$ Categorical perspective: Can the model accurately perform code translation functor?

\fparbox{\textbf{RQ3: Can the model reproduce the code based on its explanation? 
}}

(Related tasks: code explanation, code summarization, code generation, code reproduction)

$\rightarrow$ Categorical perspective: Can the model preserve the functional equivalence after applying an explanation functor and then a reproduction functor?

We conduct three experiments from categorical aspects accordingly. For a common setting, we use three multilingual datasets: HumanEval-X\cite{zheng2023codegeex}, MBXP\cite{mbxp}, and MathQA\cite{amini2019mathqa,mbxp} throughout the three experiments, 
and use Text-Davinci-003 (Text-Davinci for short), gpt3.5-turbo-0301(ChatGPT for short) as common baseline models. For more detailed experimental settings and results, please refer to Appendix~\ref{sec:impelementation_details} and ~\ref{sec:experimental_details}.

\subsection{Experiment 1: Morphism Identification Within a Code Category}

\subsubsection{Categorical Perspective Settings}
\textbf{Code Objects}
We conduct the experiments within the PL category and define ``the function with the same running outcome for every valid input'' as the same object. 

\textbf{Code Morphism}
The general category perspectives only define morphisms between two code snippets, but morphisms do not have distance information. Considering fine-grained evaluation, we are curious about the model's ability to modify or debug the code, which corresponds to the local scale; and to write an equivalent new solution to a coding problem, which is of a global scale. 

\begin{figure}[t]
    \begin{minipage}{0.49\linewidth}
    \centering
    \includegraphics[width=\linewidth]{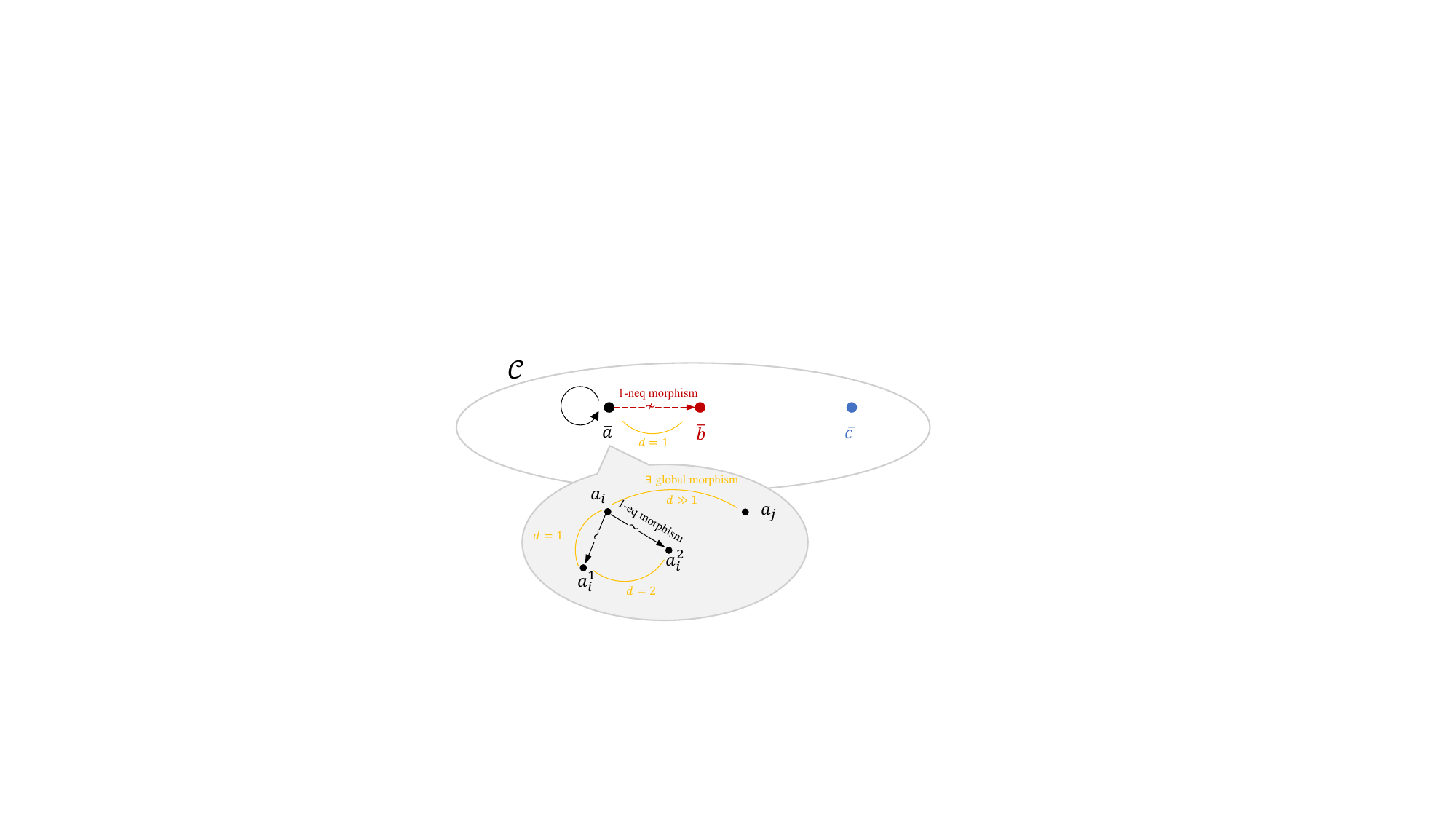}
    \end{minipage}
    \hfill
    \begin{minipage}{0.49\linewidth}
    \centering
    \includegraphics[width=\linewidth]{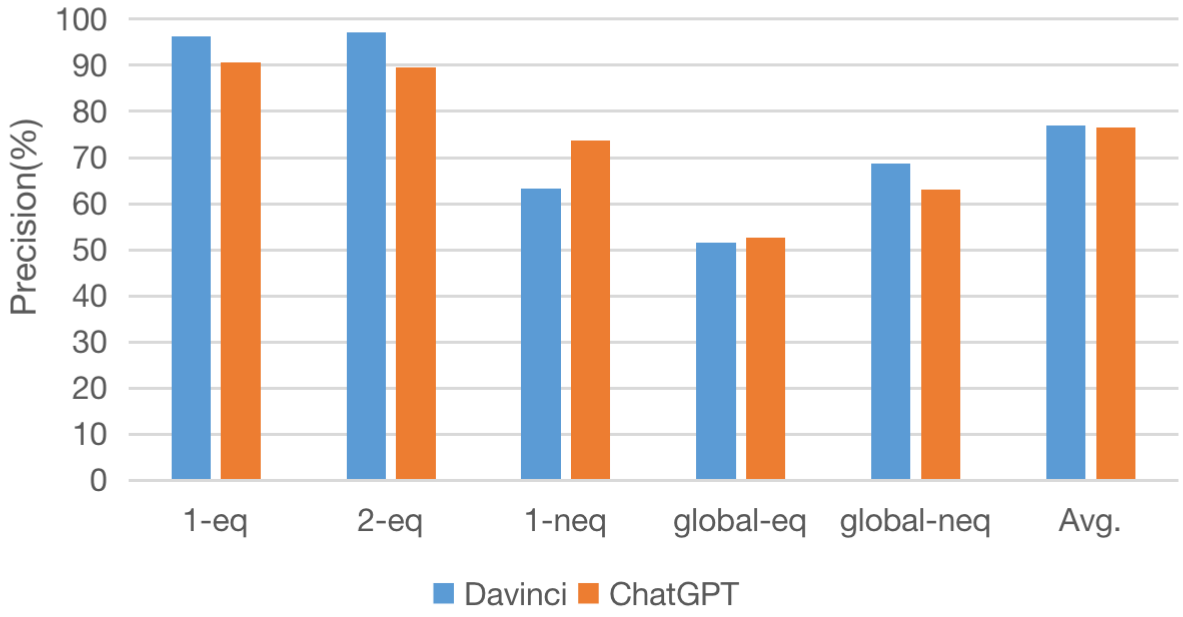}
    \end{minipage}
    \caption{Morphism Identification Experiment. ``1'', ``2'' and ``global'' stands for the distance of the code.``Eq'' and ``neq'' indicates whether the morphism is self-morphism. \textbf{(Left)} An illustration of morphisms and the definition of object distance. \textbf{(Right)} Comparison of Text-Davinci and ChatGPT for morphism identification.  }
    \label{fig:exp1}
\end{figure}

\textbf{Object Distance}
To test the model on both local and global scales, we define the ``distance'' between two codes. As illustrated in Figure \ref{fig:exp1}. In a PL Category $\mathcal{C}$, we use $\sim$ to express functional equivalence.
$\bar{a}=\{x \mid$
$x \in \mathcal{C}, x \sim a\}$ means the set of all code that has the same functionality as $a$.
In this category, $\bar{a}$ and $\bar{c}$ are two different objects. Within $\bar{a}$, all code instances are functionally equivalent, e.g. $a_i$ and $a_j$. We can apply predefined morphisms to the objects. For example, we can apply 1-step equivalent morphisms(1-eq morphism) on $a_i$ to get new program instances still in $\bar{a}$, or apply 1-step nonequivalent morphism(1-neq morphism) to get new instance in another equivalent class $\bar{b}$. Distance $d$ represents the minimal number of morphisms needed to transform one program to another given a predefined morphism set. 
For example, Applying two different self-morphism on $a_i$ returns $a_i^1$ and $a_i^2$, then we need at least two steps to transform from $a_i^1$ to $a_i^2$. Those are local-scale equivalence. If it is hard to transform $a_i$ to $a_j$ within just a few steps(\emph{e.g.}, 1 or 2 steps as illustrated here), we treat them as global-scale equivalence. 

\subsubsection{Implementation}

\textbf{Dataset} For local morphism, we extract the Java functions within HumanEval-X, MBXP, and MathQA datasets. We perform the following 9 local morphisms based on AST transformations: 

\begin{flushleft}
  1. \texttt{Variable Renaming (VR)}: rename a variable, with a new name randomly chosen from all variable names in the dataset, or use ``$var_N$'' for simplicity \\
  2. \texttt{Boolean Exchange (BE)}: propagate the exchange of ``true'' and ``false'' throughout the method and preserve the same logic \\
  3. \texttt{Loop Exchange (LE)}: exchange $for$ loops and $while$ loops \\
  4. \texttt{Switch To If (SI)}: replace a switch statement with an equivalent if statement \\
  5. \texttt{Unused Statement(US)}: insert an unused string declaration to a randomly selected basic block in a method \\
  6. \texttt{Reorder Condition (RC)}: write the condition in reverse order (\emph{e.g.}, change $i<n$ to $n>i$) \\
  7. \texttt{Permute Statement(PS)}: swap two independent statements (i.e. without data or control dependence) \\
  8. \texttt{Modify Condition(MC)}: change the binary operation in the condition (\emph{e.g.}, change $i<n$ to $i\leq n$) \\
  9. \texttt{Remove Else(RE)}: delete the else branch of the if-else statement \\
\end{flushleft}

The first 7 morphisms are adopted from JavaTransform\cite{javatransformer} that conducts functionally equivalent transformations \footnote{Transformation samples can refer to: https://github.com/mdrafiqulrabin/tnpa-framework .}, which means they are self-morphisms. Other than self-morphisms, we add 2 morphisms: \texttt{ModifyCondition} and  \texttt{RemoveElse}, that change the program's function.

For global morphisms, since the three datasets above do not contain multiple solutions to the same problem using the same PL, we complement with code from the test split of Code Contest\cite{alphacode2022} dataset.

\textbf{Models} We evaluate Text-Davinci and ChatGPT. We input (code, code) pairs from the datasets or generated by morphisms, and ask the model to answer whether they are functionally equivalent. 

\textbf{Evaluation} We collect the model's responses and calculate the average precision for different morphisms.

\subsubsection{Results} 
The results are demonstrated in Figure \ref{fig:exp1}. Overall, Text-Davinci and ChatGPT make a tie. Locally, ChatGPT is better at identifying nonequivalent morphisms, and worse at equivalent ones. Globally, Text-Davinci is better at identifying global nonequivalent morphisms.  Note that a random guess leads to a precision of 50\%, so both models behave just slightly better than random guesses for identifying global equivalence.

\subsection{Experiment 2: Translation Functor Between Different PL Categories}

\subsubsection{Categorical Perspective Settings}
We investigate the model's ability to perform code translation by utilizing functors between two programming language categories. To isolate the influence of natural language, we exclude problem descriptions and code comments, forcing the models to focus solely on translating code from one programming language (PL) category to another.
The input prompt may contain a request for translation using natural language (\emph{e.g.}, ``translate the below Java code to Python code'').

\subsubsection{Implementation}

\textbf{Datasets} We conduct evaluations on the HumanEval-X, MathQA, and MBXP datasets. The input category is Java, while the output categories are Python and JavaScript.

\textbf{Models} We evaluate three models: Text-Davinci, ChatGPT, and CodeGeeX. For the first two models, we provide prompts that request the model to do the translation. However, for CodeGeeX, code translation is supported inherently, so we simply input the Java code without additional text.

\textbf{Evaluation} We extract the functions from the model's responses and assess their correctness using the Pass@1 rates of execution-based tests.

\subsubsection{Results} 
\begin{figure}[t]
    \begin{minipage}{0.50\linewidth}
        \centering
        \includegraphics[width=\linewidth]{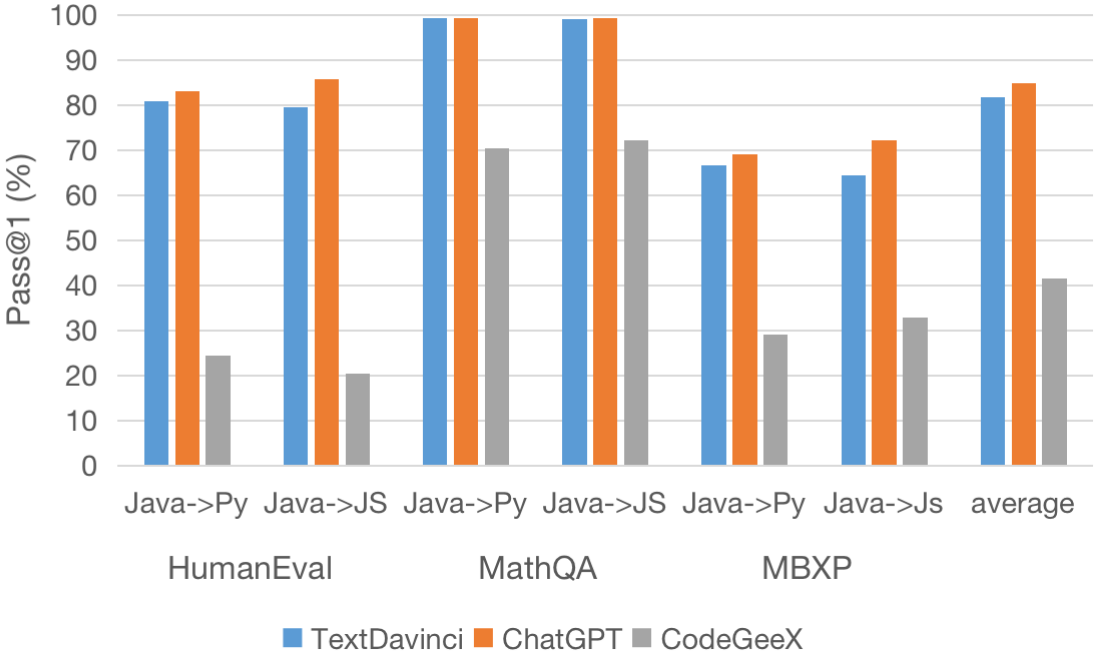}
    \end{minipage}
    \hfill
    \begin{minipage}{0.48\linewidth}
        \centering
        \includegraphics[width=\linewidth]{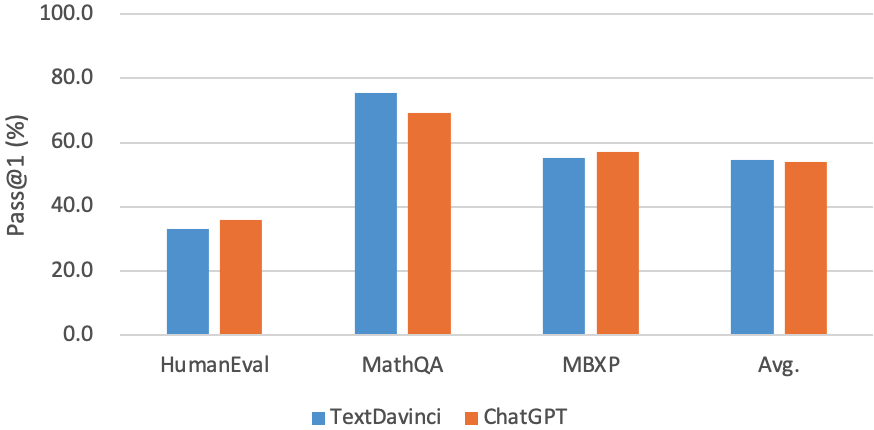}
    \end{minipage}
    \caption{Comparison of model performance. \textbf{(Left)} Model as a translation functor. \textbf{(Right)} Model as the combination of explanation functor and reproduction functor}
    \label{fig:exp2-3}
\end{figure}

The results are depicted in Figure \ref{fig:exp2-3}. Among the three models, ChatGPT performs the best, exhibiting a slight advantage over Text-Davinci. With regards to the datasets, ChatGPT and Text-Davinci achieve nearly perfect translation accuracy on MathQA, whose data consists of functions with internal variable initialization, and without any input arguments, loops, or conditional statements. This indicates ChatGPT's and Text-Davinci's proficiency in reproducing exact numerical values and handling simple program structures.

\subsection{Experiment 3: Explanation Functor and Reproduction Functor Between PL and NL Categories}

\subsubsection{Categorical Perspective Settings}
Code explanation and code generation tasks have been conducted separately in previous work. However, from a categorical perspective, we can make a combination. We define the explanation functor as the functor that maps an object from a PL category to an NL category, with a precise description of the code's function. We define the reproduction functor as the functor from an NL category to a PL category, that uses an explanation object in NL to reproduce the code of the same function in a PL category. If the reproduced code is functionally the same as the original code, we reckon the model preserves the functional equivalence after applying an explanation functor and then a generation functor.

\subsubsection{Implementation}

\textbf{Datasets} We conduct the evalution on three datasets: HumanEval-X, MathQA, and MBXP. For consistency, we select Java as the programming language (PL) category for all evaluations.

\textbf{Models} We evaluate Text-Davinci and ChatGPT, prompting each model twice. In the first prompt, we ask the models to describe the code using natural language, including the precise function name, arguments, and return type, and provide sufficient information to reproduce the code. In the second prompt, we provide the model's explanation and ask it to translate the description back into code.

\textbf{Evaluation} To evaluate the correctness of the reproduced code, we extract the functions from the model's second responses and assess them using the Pass@1 rates of execution-based tests. 

\subsubsection{Results} 
The results are depicted in Figure \ref{fig:exp2-3}. On the same datasets and with the same two models, the average pass@1 rate is about 30\% lower compared to that of the translation functor. This indicates the model has significant information loss in the explain-reproduce process.

\subsection{Summary}
Based on our experiments, we have found that both Text-Davinci and ChatGPT models are capable of identifying the impact of one or two local morphisms on code function. However, identifying global morphisms proves to be challenging for these models. In terms of code translation, both ChatGPT and Text-Davinci models outperform CodeGeex. They demonstrate accurate reproduction of basic syntax, variable names, and numerical values. Nevertheless, when dealing with more intricate program structures and aligning data types across different programming language categories, these models encounter obstacles that hinder more precise translation. It is important to note that maintaining functional equivalence between code and natural language presents a more formidable challenge compared to code translation, necessitating further research in this specific domain.

\section{Related Work}
\subsection{Coding Abilities of Large Language Models}
Large language models trained on code have demonstrated improvements in various coding tasks. CodeBERT \cite{codebert} was one of the initial models trained on code, while GraphCodeBERT \cite{guo2020graphcodebert} incorporated program Abstract Syntax Trees (AST) and data flow information to enhance code structure and context understanding.

The Encoder-Decoder architecture, used in models like CodeT5 \cite{CodeT5} and PLBART \cite{PLBART}, enables multi-task learning for code translation, understanding, and generation. PLBART\cite{PLBART}, pretrained on Java and Python functions with natural language comments, excels in generating code from natural language descriptions.

More recently, the emergence of Codex \cite{codex} and ChatGPT\footnote{https://openai.com/blog/chatgpt/} has brought GPT-based models into the mainstream for content generation tasks, including coding. These models have shown impressive performance across various domains, but their proprietary nature, limited access to data, and lack of model checkpoints have made it challenging to thoroughly evaluate and understand their capabilities and limitations. Consequently, there is a need to develop a comprehensive framework to evaluate these black-box models.

\subsection{Code Model Evaluation}
There are four widely-used types of evaluation frameworks for code models: match-based, attack-based, task-based, and execution-based.

The match-based framework includes CodeBLEU\cite{codebleu2020} that adopts n-gram match as BLEU\cite{bleu} and further injects Abstract Syntax Tree(AST) and data-flow similarity. 
However, match-based framework may consider different solutions with varying variable names, AST structures, and data flow as dissimilar, despite their functional equivalence.

The attack-based framework constructs adversarial program examples to assess the model's performance. Yefet et al. \cite{Yefet} generated adversarial code examples based on gradients, uncovering vulnerabilities to variable renaming and dead code injection.  
Similarly, Ramakrishnan et al.\cite{averloc-HenkelRWAJR22} and Rabin et al.\cite{semantic-preserving} investigated the model's robustness and generalizability to semantic-preserving program transformations.
However, those works primarily focused on white-box models, and the adversarial examples are not so commonly encountered in practice. Nevertheless, the idea of ``functional equivalence'' and ``program transformations'' in their framework is important for related research.

Task-based evaluations focus on the model's performance on specific coding tasks. CodeXGLUE\cite{lu2021codexglue} is a well-known task-based framework that employs diverse datasets for different coding tasks. 
It provides a very comprehensive evaluation framework. However, different tasks in it do not share enough common settings in data definition, task formulation, and test APIs. This hinders its use in a more scalable and easy-to-follow way.

The execution-based evaluation focuses on code generation and translation tasks, evaluating the correctness of the model-generated code based on test cases. MBXP\cite{mbxp} and MultiPL-E\cite{cassano_multipl-e_2023} are two execution-based benchmarks that support multiple programming languages. While our work also considers test cases from execution-based benchmarks, our focus extends beyond correctness evaluation to include code morphism and PL to NL functor aspects.

\section{Limitations}

While we strive to apply categorical perspectives to offer a comprehensive and standardized way of evaluation, we find that our current study has a number of limitations. First, more powerful concepts and properties in category theory haven't been fully utilized by our current study. Our experiments primarily focus on objects, morphisms, functors, and their combinations. However, we leave it for future work to further investigate the setting of composing many morphisms and many functors to further explore the commutative law and isomorphism properties of categories. Second, we use prompts to instruct the models to act as certain functors, but prompts that express similar meanings will lead to different results, we have not investigated which prompts are more suitable for the model or which models are more robust to different prompts. Moreover, due to limitations in computational resources and API accessibilities, we did not test more models. 

\section{Conclusion}

Our contributions include introducing CatCode as a novel evaluation perspective based on category theory, which enables a comprehensive and mathematically abstract approach to evaluate LLMs that excel in understanding and generating a mixture of code and text. Based on categorical perspectives, we present a standardized automatic evaluation platform, which is adaptive to new datasets, tasks, and models. We evaluate competitive LLMs to provide valuable insights, and find out current models' deficiency in recognizing functionally equivalent code and preserving information of code function between the code and its explanation.
We plan to open-source our platform, hoping to contribute to the comprehensive and standardized evaluation for LLMs by offering a categorical perspective to deal with the mixture of code and text.

We believe that CatCode represents an important step towards a more comprehensive and standardized evaluation of LLMs' abilities in solving coding problems. By combining the power of category theory with the growing capabilities of LLMs, we can unlock new possibilities for defining and evaluating more diverse code-related tasks. We hope that CatCode inspires further research and development in the field, leading to more sophisticated LLMs that can effectively assist developers in their coding tasks and contribute to advancements in artificial intelligence and software engineering as a whole.

\newpage

\bibliography{catcode_arxiv}{}

\begin{thebibliography}{10}

\bibitem{PLBART}
Wasi~Uddin Ahmad, Saikat Chakraborty, Baishakhi Ray, and Kai{-}Wei Chang.
\newblock Unified pre-training for program understanding and generation.
\newblock In Kristina Toutanova, Anna Rumshisky, Luke Zettlemoyer, Dilek
  Hakkani{-}T{\"{u}}r, Iz~Beltagy, Steven Bethard, Ryan Cotterell, Tanmoy
  Chakraborty, and Yichao Zhou, editors, {\em Proceedings of the 2021
  Conference of the North American Chapter of the Association for Computational
  Linguistics: Human Language Technologies, {NAACL-HLT} 2021, Online, June
  6-11, 2021}, pages 2655--2668. Association for Computational Linguistics,
  2021.

\bibitem{functionalprogramming}
Benedikt Ahrens and Kobe Wullaert.
\newblock Category theory for programming.
\newblock {\em CoRR}, abs/2209.01259, 2022.

\bibitem{amini2019mathqa}
Aida Amini, Saadia Gabriel, Peter Lin, Rik Koncel-Kedziorski, Yejin Choi, and
  Hannaneh Hajishirzi.
\newblock Mathqa: Towards interpretable math word problem solving with
  operation-based formalisms.
\newblock {\em arXiv preprint arXiv:1905.13319}, 2019.

\bibitem{mbxp}
Ben Athiwaratkun, Sanjay~Krishna Gouda, Zijian Wang, Xiaopeng Li, Yuchen Tian,
  Ming Tan, Wasi~Uddin Ahmad, Shiqi Wang, Qing Sun, Mingyue Shang, Sujan~Kumar
  Gonugondla, Hantian Ding, Varun Kumar, Nathan Fulton, Arash Farahani,
  Siddhartha Jain, Robert Giaquinto, Haifeng Qian, Murali~Krishna Ramanathan,
  Ramesh Nallapati, Baishakhi Ray, Parminder Bhatia, Sudipta Sengupta, Dan
  Roth, and Bing Xiang.
\newblock Multi-lingual evaluation of code generation models.
\newblock {\em CoRR}, abs/2210.14868, 2022.

\bibitem{bradley2018applied}
Tai-Danae Bradley.
\newblock What is applied category theory?
\newblock {\em arXiv preprint arXiv:1809.05923}, 2018.

\bibitem{cassano_multipl-e_2023}
Federico Cassano, John Gouwar, Daniel Nguyen, Sydney Nguyen, Luna
  Phipps-Costin, Donald Pinckney, Ming-Ho Yee, Yangtian Zi, Carolyn~Jane
  Anderson, Molly~Q Feldman, Arjun Guha, Michael Greenberg, and Abhinav Jangda.
\newblock {MultiPL}-{E}: {A} {Scalable} and {Polyglot} {Approach} to
  {Benchmarking} {Neural} {Code} {Generation}.
\newblock {\em IEEE Transactions on Software Engineering}, pages 1--17, 2023.

\bibitem{codex}
Mark Chen, Jerry Tworek, Heewoo Jun, Qiming Yuan, Henrique~Ponde
  de~Oliveira~Pinto, Jared Kaplan, Harrison Edwards, Yuri Burda, Nicholas
  Joseph, Greg Brockman, Alex Ray, Raul Puri, Gretchen Krueger, Michael Petrov,
  Heidy Khlaaf, Girish Sastry, Pamela Mishkin, Brooke Chan, Scott Gray, Nick
  Ryder, Mikhail Pavlov, Alethea Power, Lukasz Kaiser, Mohammad Bavarian,
  Clemens Winter, Philippe Tillet, Felipe~Petroski Such, Dave Cummings,
  Matthias Plappert, Fotios Chantzis, Elizabeth Barnes, Ariel Herbert-Voss,
  William~Hebgen Guss, Alex Nichol, Alex Paino, Nikolas Tezak, Jie Tang, Igor
  Babuschkin, Suchir Balaji, Shantanu Jain, William Saunders, Christopher
  Hesse, Andrew~N. Carr, Jan Leike, Joshua Achiam, Vedant Misra, Evan Morikawa,
  Alec Radford, Matthew Knight, Miles Brundage, Mira Murati, Katie Mayer, Peter
  Welinder, Bob McGrew, Dario Amodei, Sam McCandlish, Ilya Sutskever, and
  Wojciech Zaremba.
\newblock Evaluating large language models trained on code.
\newblock {\em CoRR}, abs/2107.03374, 2021.

\bibitem{codebert}
Zhangyin Feng, Daya Guo, Duyu Tang, Nan Duan, Xiaocheng Feng, Ming Gong, Linjun
  Shou, Bing Qin, Ting Liu, Daxin Jiang, et~al.
\newblock Codebert: A pre-trained model for programming and natural languages.
\newblock {\em arXiv preprint arXiv:2002.08155}, 2020.

\bibitem{fong2018seven}
Brendan Fong and David~I Spivak.
\newblock Seven sketches in compositionality: An invitation to applied category
  theory.
\newblock {\em arXiv preprint arXiv:1803.05316}, 2018.

\bibitem{guo2020graphcodebert}
Daya Guo, Shuo Ren, Shuai Lu, Zhangyin Feng, Duyu Tang, Shujie Liu, Long Zhou,
  Nan Duan, Alexey Svyatkovskiy, Shengyu Fu, et~al.
\newblock Graphcodebert: Pre-training code representations with data flow.
\newblock {\em arXiv preprint arXiv:2009.08366}, 2020.

\bibitem{averloc-HenkelRWAJR22}
Jordan Henkel, Goutham Ramakrishnan, Zi~Wang, Aws Albarghouthi, Somesh Jha, and
  Thomas~W. Reps.
\newblock Semantic robustness of models of source code.
\newblock In {\em {IEEE} International Conference on Software Analysis,
  Evolution and Reengineering, {SANER} 2022, Honolulu, HI, USA, March 15-18,
  2022}, pages 526--537. {IEEE}, 2022.

\bibitem{alphacode2022}
Yujia Li, David Choi, Junyoung Chung, Nate Kushman, Julian Schrittwieser, Rémi
  Leblond, Tom Eccles, James Keeling, Felix Gimeno, Agustin~Dal Lago, Thomas
  Hubert, Peter Choy, Cyprien de~Masson~d’Autume, Igor Babuschkin, Xinyun
  Chen, Po-Sen Huang, Johannes Welbl, Sven Gowal, Alexey Cherepanov, James
  Molloy, Daniel~J. Mankowitz, Esme~Sutherland Robson, Pushmeet Kohli, Nando
  de~Freitas, Koray Kavukcuoglu, and Oriol Vinyals.
\newblock Competition-level code generation with alphacode.
\newblock {\em Science}, 378(6624):1092--1097, 2022.

\bibitem{lu2021codexglue}
Shuai Lu, Daya Guo, Shuo Ren, Junjie Huang, Alexey Svyatkovskiy, Ambrosio
  Blanco, Colin Clement, Dawn Drain, Daxin Jiang, Duyu Tang, et~al.
\newblock Codexglue: A machine learning benchmark dataset for code
  understanding and generation.
\newblock {\em arXiv preprint arXiv:2102.04664}, 2021.

\bibitem{gpt3explain-MacNeilTMBRH22}
Stephen MacNeil, Andrew Tran, Dan Mogil, Seth Bernstein, Erin Ross, and Ziheng
  Huang.
\newblock Generating diverse code explanations using the {GPT-3} large language
  model.
\newblock In Jan Vahrenhold, Kathi Fisler, Matthias Hauswirth, and Diana
  Franklin, editors, {\em {ICER} 2022: {ACM} Conference on International
  Computing Education Research, Lugano and Virtual Event Switzerland, August 7
  - 11, 2022, Volume 2}, pages 37--39. {ACM}, 2022.

\bibitem{instructgpt}
Long Ouyang, Jeffrey Wu, Xu~Jiang, Diogo Almeida, Carroll~L. Wainwright, Pamela
  Mishkin, Chong Zhang, Sandhini Agarwal, Katarina Slama, Alex Ray, John
  Schulman, Jacob Hilton, Fraser Kelton, Luke Miller, Maddie Simens, Amanda
  Askell, Peter Welinder, Paul~F. Christiano, Jan Leike, and Ryan Lowe.
\newblock Training language models to follow instructions with human feedback.
\newblock In {\em NeurIPS}, 2022.

\bibitem{bleu}
Kishore Papineni, Salim Roukos, Todd Ward, and Wei{-}Jing Zhu.
\newblock Bleu: a method for automatic evaluation of machine translation.
\newblock In {\em Proceedings of the 40th Annual Meeting of the Association for
  Computational Linguistics, July 6-12, 2002, Philadelphia, PA, {USA}}, pages
  311--318. {ACL}, 2002.

\bibitem{semantic-preserving}
Md. Rafiqul~Islam Rabin, Nghi D.~Q. Bui, Ke~Wang, Yijun Yu, Lingxiao Jiang, and
  Mohammad~Amin Alipour.
\newblock On the generalizability of neural program models with respect to
  semantic-preserving program transformations.
\newblock {\em Inf. Softw. Technol.}, 135:106552, 2021.

\bibitem{javatransformer}
Md. Rafiqul~Islam Rabin, Ke~Wang, and Mohammad~Amin Alipour.
\newblock Testing neural programs.
\newblock {\em CoRR}, abs/1908.10711, 2019.

\bibitem{codebleu2020}
Shuo Ren, Daya Guo, Shuai Lu, Long Zhou, Shujie Liu, Duyu Tang, Neel
  Sundaresan, Ming Zhou, Ambrosio Blanco, and Shuai Ma.
\newblock Codebleu: a method for automatic evaluation of code synthesis.
\newblock {\em CoRR}, abs/2009.10297, 2020.

\bibitem{cat_science}
David~I. Spivak.
\newblock {\em Category Theory for the Sciences}.
\newblock {MIT} Press, 2014.

\bibitem{intelliCode2020Svyatkovskiy}
Alexey Svyatkovskiy, Shao~Kun Deng, Shengyu Fu, and Neel Sundaresan.
\newblock Intellicode compose: code generation using transformer.
\newblock In Prem Devanbu, Myra~B. Cohen, and Thomas Zimmermann, editors, {\em
  {ESEC/FSE} '20: 28th {ACM} Joint European Software Engineering Conference and
  Symposium on the Foundations of Software Engineering, Virtual Event, USA,
  November 8-13, 2020}, pages 1433--1443. {ACM}, 2020.

\bibitem{CodeT5}
Yue Wang, Weishi Wang, Shafiq~R. Joty, and Steven C.~H. Hoi.
\newblock Codet5: Identifier-aware unified pre-trained encoder-decoder models
  for code understanding and generation.
\newblock In Marie{-}Francine Moens, Xuanjing Huang, Lucia Specia, and
  Scott~Wen{-}tau Yih, editors, {\em Proceedings of the 2021 Conference on
  Empirical Methods in Natural Language Processing, {EMNLP} 2021, Virtual Event
  / Punta Cana, Dominican Republic, 7-11 November, 2021}, pages 8696--8708.
  Association for Computational Linguistics, 2021.

\bibitem{Yefet}
Noam Yefet, Uri Alon, and Eran Yahav.
\newblock Adversarial examples for models of code.
\newblock {\em Proc. {ACM} Program. Lang.}, 4({OOPSLA}):162:1--162:30, 2020.

\bibitem{zheng2023codegeex}
Qinkai Zheng, Xiao Xia, Xu~Zou, Yuxiao Dong, Shan Wang, Yufei Xue, Zihan Wang,
  Lei Shen, Andi Wang, Yang Li, Teng Su, Zhilin Yang, and Jie Tang.
\newblock Codegeex: A pre-trained model for code generation with multilingual
  evaluations on humaneval-x, 2023.

\bibitem{zhu2022multilingual}
Ming Zhu, Karthik Suresh, and Chandan~K Reddy.
\newblock Multilingual code snippets training for program translation.
\newblock In {\em Proceedings of the AAAI Conference on Artificial
  Intelligence}, volume~36, pages 11783--11790, 2022.

\end{thebibliography}
\bibliographystyle{plain}

\newpage

\appendix

\begin{center}
    \huge{Appendix}
\end{center}

The appendix is divided into three main sections, each focusing on distinct aspects of the study. These sections are labeled as follows:

A. Implementation Details: This section covers the implementation process, including the anonymous link to our code, dataset details, data filtering, model specifications, and evaluation metrics, providing information on how to reproduce based on our standard evaluation pipeline.

B. Experimental Details: This part introduces the experimental settings and more experimental results in detail. We further discuss the model's performance on different morphism types in the identification experiment, and the common error types in the translation and reproduction experiments.

C. Case Study: This section offers some intuitive examples of the experiments. Through analysis of those examples, we can find some limitations of current models, prompts, and test scripts.

\section{Implementation Details}
\label{sec:impelementation_details}

We have uploaded the code to the anonymous website: https://anonymous.4open.science/r/CatCode-6402/. A \texttt{README.md} file is provided, offering step-by-step instructions on how to use the evaluation platform and replicate the entire evaluation process in a standardized way.

In this section, we will provide additional details regarding the reproduction process, including the dataset split, data filtering, models used, and evaluation metrics.

\subsection{Dataset Details}
\label{sec:appendix_dataset_details}

The following datasets were used in our evaluation:

\begin{itemize}[leftmargin=*]
    \item HumanEval-X\footnote{https://huggingface.co/datasets/THUDM/humaneval-x}: a human-crafted dataset with 164 problems, correct solutions in 5 different languages, and test cases. It is a benchmark for evaluating code generation and translation models. We use the Java split of HumanEval-X as input for applying morphism, code translation, and code explanation. Although \texttt{Mxeval} provide multilingual solutions, they are generated by models and do not make a distinction between correct and wrong solutions. In contrast, HumanEval-X provides correct solutions, making it a more suitable complement.
    \item MBXP: a benchmark for evaluating code generation models. It covers over 10 programming languages and is generated using a scalable conversion framework that transpiles prompts and test cases from the original Python datasets into the corresponding data in the target language. It's a dataset similar to HumanEval-X but does not ensure the correctness of the transpiled code. 
    \item MathQA: a dataset for evaluating math problem solvers. All functions in this dataset have no input argument and return a number. The functions first initialize variables, then do some calculations, and do not use any loops or conditional statements. Compared to HumanEval-X and MBXP, the code in MathQA is simpler in structure and functionality.
    \item Code Contest: a large-scale competitive multi-language programming dataset used for training AlphaCode\cite{alphacode2022}. It consists of programming problems, test cases in the form of paired inputs and outputs, and both correct and incorrect solutions. Since the aforementioned datasets do not provide multiple solutions within the same programming language, Code Contest serves as a valuable complement by offering multiple solutions to the same problem, representing global equivalence and non-equivalence.
\end{itemize}

\textbf{Data filtering}
\begin{itemize}[leftmargin=*]
    \item Only the Java split of HumanEval-X, MBXP, and MathQA datasets are used. The local morphism is applied by \texttt{JavaTransformer} by AST edits at the function level. To use JavaTransformer, code must compile successfully and contain a single function. We filter out code in the \texttt{Mxeval} datasets that are generated by the model and cannot compile or contain multiple functions.
    \item Due to the large size of the Code Contest dataset, only the test set is used. Considering that the maximum token length of Text-Davinci is 2048, we filter out solutions with a length greater than 500. This ensures that the code pair length remains $\leq$ 1000, allowing for additional tokens for prompt content and response. After filtering, 97 problems remain, and a problem may have multiple correct and incorrect solutions in different languages. We utilize the Java and Python splits of this dataset.
\end{itemize}

\subsection{Model details}
We utilize the following baseline models by making API calls to their official endpoints: Text-Davinci-003 (referred to as Davinci), ChatGPT, and CodeGeeX. The default hyperparameters of these models are used, with the exception of setting \texttt{max\_token} to 500 when an early stop of the answer is observed. 

Table \ref{tab:baselines} provides an overview of the tasks supported by each model. Since Text-Davinci-003 and ChatGPT support various types of text inputs and outputs, they are capable of performing all tasks. On the other hand, the CodeGeeX API does not support natural language output, so we only employ it in translation functor experiments.

\begin{table}[t]
    \centering
    \begin{tabular}{ccccc}
    \hline
       Model & Morphism& Translation Functor &  Explanation Functor & Reproduction Functor\\ 
    \hline
       
        Davinci  & \checkmark& \checkmark& \checkmark& \checkmark \\
        ChatGPT  & \checkmark& \checkmark& \checkmark& \checkmark \\
        CodeGeeX &           & \checkmark&           &  \\
       
    \hline
    \end{tabular}
    \caption{Baseline Models and their supported evaluation tasks. }
    \label{tab:baselines}
\end{table}

\subsection{Evaluation Metrics}

To assess the performance of the model, we conduct two types of tests: pairing test and execution-based test.

The pairing test is utilized for morphism identification. The task involves comparing two code snippets and requesting the model to explicitly answer ``True'' or ``False'', along with providing a comparison. For automatic evaluation, we extract the        True'' or ``False'' answer. The precision score is used to evaluate the model's ability to identify different equivalence classes. For equivalent objects, the precision is calculated as $Precision(eq)=\frac{TP}{TP+FN}$, while for nonequivalent objects, the precision is calculated as $Precision(neq)=\frac{TN}{TN+FP}$.

The execution-based test is employed for both translation and reproduction experiments. We extract the functions from the model's responses and evaluate their correctness using the Pass@1 rates of execution-based tests. This test ensures that the translated/reproduced code produces the same expected results as the original code.

\section{Experimental Details}
\label{sec:experimental_details}

\subsection{Morphism Identification}

\textbf{Data statistics}.
Table~\ref{tab:morphism_statistics} provides the statistics of the raw data, filtered data, and constructed pairs for the morphism identification experiment. 
The filtering strategy, described in Section~\ref{sec:appendix_dataset_details}, involves removing comments and docstrings, resulting in filtered data consisting solely of Java code.
The constructed pairs are generated through sampling. 
It is worth noting that certain morphisms can have multiple outputs for a given input. For instance, the \texttt{Variable Renaming} morphism can be applied to all variables in a function, resulting in multiple outputs equal to the number of variables. To balance the number of outputs across different morphism types, random sampling is employed when there are more than two applicable morphism types. For each code sample, two types of morphisms are first sampled, followed by sampling one output for each morphism type. 

\begin{table}[t]
  \begin{center}
    \begin{tabular}{cccc}
    \hline
      \textbf{Dataset} & \textbf{\#after filter / \#raw data} & \textbf{\#equivalent pair}& \textbf{\#nonequivalent pair}\\
      \hline
      Humaneval-X & 159 / 164 & 477 & 173\\
      MBXP & 953 / 974 & 2849 & 1010\\
      MathQA & 1734 / 1881 & 4852 & 1418\\
      Code Contest & 97 / 164 & 97 & 366 \\
    \hline
    \end{tabular}
    \caption{Dataset statistics for morphism identification experiment.}
    \label{tab:morphism_statistics}
  \end{center}
\end{table}

\textbf{Discussion of different morphism types.}

Which morphism types are harder to be identified, and are they hard  across datasets? We calculate the average precision scores for 1-eq, 2-eq and 1-neq local morphisms for ChatGPT answers, as shown in Table~\ref{tab:morphism_type}. 

The following observations can be made:

\begin{itemize}[leftmargin=*]
    \item For one morphism: ``Unused Statements'', ``Modify Condition'' and ``Boolean Exchange'' are particularly difficult to identify. One possible reason for this difficulty is that these morphisms involve subtle changes or transformations in the code that may not be easily recognizable based on the surrounding context alone. It requires a more nuanced understanding of the code logic and structure to detect these morphisms 
    \item For 2-eq morphisms: The observation that ``Unused Statements'' remains challenging when combined with other morphisms suggests that the presence of multiple morphisms in the code can further complicate the identification task. 
    \item Dataset-specific difficulties: The results indicate that the difficulty of identifying ``Unused Statements'' persists across datasets, suggesting that this morphism type poses inherent challenges in code comprehension. On the other hand, the specific difficulty in identifying ``Modify Condition'' in the MathQA dataset may be attributed to the nature of the dataset itself, which focuses on math problem solvers.
\end{itemize}

\begin{table}[t]
    \centering
    \begin{tabular}{llll}
    \hline
         HumanEval & MathQA & MBXP \\ \hline
          BE-VR (66.67) & MC (45.66) & BE (33.33)\\ 
          US (74.74) & RC-US (85.71) & PS-US (69.57) \\ 
          LE-RC (78.57) & PS-US (87.84) & RC-US (70.39) \\ \hline
    \end{tabular}
    \caption{Morphism types with the three worst average precision scores of each dataset. As an example for notations, BE-VR (66.67) stands for a 2-eq morphism of ``Boolean Exchange'' and ``Variable Renaming'' with a precision score of 66.67\%. }
    \label{tab:morphism_type}
\end{table}

\subsection{Translation Functor}
\label{appendix:translation_functor}

\textbf{Data statistics}.
We use the filtered data in Table~\ref{tab:morphism_statistics} as input, i.e. 159, 953, and 97 Java Snippets for HumanEval-X, MBXP, and MathQA, respectively. 
Each model generates a Python object and a JavaScript object for each Java object. 
During test phase, since the $\texttt{Mexeval}$ dataset is missing 3 test cases in the Java split of HumanEval (TaskID: 32, 38, 50), we use 156, 953, and 97 Java functions to calculate the pass@1 rates for HumanEval-X, MBXP, and MathQA respectively. 

\textbf{Results statistics}. The pass@1 score of different models on different datasets is shown in Table~\ref{fig:translation_table}. Overall, ChatGPT has a relatively higher translation ability.

\begin{table}[t]
    \centering
    \includegraphics[width=\linewidth]{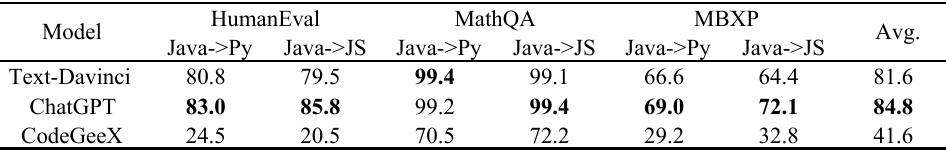}
    \caption{Translation pass@1 score (\%) statistics, corresponding to the Figure~\ref{fig:exp2-3} \textbf{(Left)}. The models translate Java code to two languages Python (Py) and JavaScript (JS), on three datasets. The average (Avg.) is the average score over the six scores.}
    \label{fig:translation_table}
\end{table}

\textbf{Discussion of failure types.} Table~\ref{fig:translation_types} represents the most frequent failure types of translation functor on three datasets. We observe that most errors are compilation errors rather than failures at corner test cases. 

These errors are often caused by type mismatches between different programming languages. For instance, when translating a list in Java, there are multiple potential types in Python, including list, tuple, and List from the typing module. 

The occurrence of ``NameError'' and ``ReferenceError'' indicates an unfaithful translation of variable names by the model or mismatched function input arguments across different languages for the same problem. This is a limitation inherent in the current evaluation setup.

Overall, the models have demonstrated the ability to perform correct translations in most cases between different programming languages. 
The errors mainly arise from type alignment issues between programming languages and local variable definitions. These errors are typically easy to debug for programmers if provided with relevant information. Future work could involve passing error messages to the model and allowing it to debug itself.

\begin{table}[t]
    \centering
    \includegraphics[width=\linewidth]{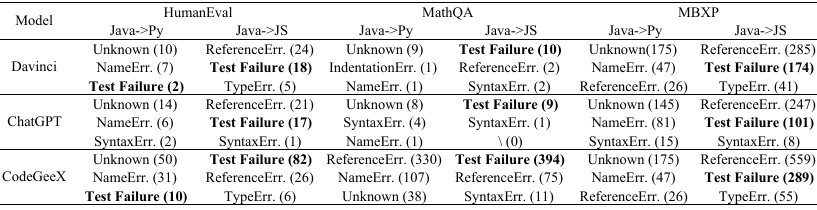}
    \caption{Most frequent failure types of translation functor. The data in the table cell is written as ErrorType (frequency), and Err. is short for Error. Specifically, the ``Unknown'' type only exists in Python and ``Test Failure'' represents the code that complies successfully but fails to pass some of the test cases. Other types of errors in this table are primarily caused by compilation errors. }
    \label{fig:translation_types}
\end{table}

\subsection{Explanation and Reproduction Functor}

\textbf{Data statistics}.
Same as Appendix~\ref{appendix:translation_functor}, we use 156, 953, and 97 Java functions to calculate the pass@1 rates for HumanEval-X, MBXP, and MathQA separately. 

\textbf{Results statistics}. Table~\ref{fig:reproduction_table} shows the pass@1 rates of different models. TextDavinci is better at MathQA, while ChatGPT is better at HumanEval and MBXP datasets. The pass@1 score is significantly lower compared to translation functor experiments, indicating that it's a more difficult task. 

\begin{table}[t]
    \centering
    \begin{tabular}{ccccc}
    \hline
        Model & HumanEval & MathQA & MBXP & Avg. \\ \hline
        TextDavinci & 33.0  & \textbf{75.5}  & 55.2  & \textbf{54.6}  \\ 
        ChatGPT & \textbf{35.9}  & 69.1  & \textbf{57.1}  & 54.0 \\ \hline
    \end{tabular}
    \caption{Pass@1 score (\%) statistics of the reproduced code, corresponding to the Figure~\ref{fig:exp2-3} \textbf{(Right)}.}
    \label{fig:reproduction_table}
\end{table}

\textbf{Discussion of failure types.} Table~\ref{fig:reproduction_types} shows the most frequent failure types of translation functor on the three datasets. Based on the results, it appears that both models encounter similar types of failures across the different evaluation tasks. The ``Test Failure'' type is particularly prominent, indicating that the reproduced code, although compiling successfully, fails to pass some of the test cases. Additionally, both models also encounter errors related to symbol identification, such as ``CannotFindSymbol'' and ``TypeError''. These errors are primarily compilation errors.

It's worth further investigating the causes behind these failure types and exploring strategies to improve the models' performance, especially in handling test cases and resolving symbol-related errors.

\begin{table}[t]
    \centering
    \includegraphics[width=0.8\linewidth]{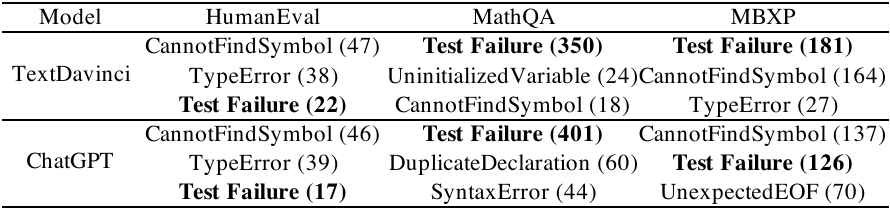}
    \caption{Most frequent failure types of the reproduced code. The data in the table cell is written as ErrorType(frequency). Specifically, the ``Test Failure'' type represents the code that complies successfully but fails to pass some of the test cases. Other types of errors in this table are primarily caused by compilation errors. }
    \label{fig:reproduction_types}
\end{table}

\section{Case Study}
\label{sec:case_study}
In this Section, we analyze some classical cases from our three experiments.

\subsection{Morphism Identification}

\begin{figure}[t]
    \centering
    \includegraphics[width=\linewidth]{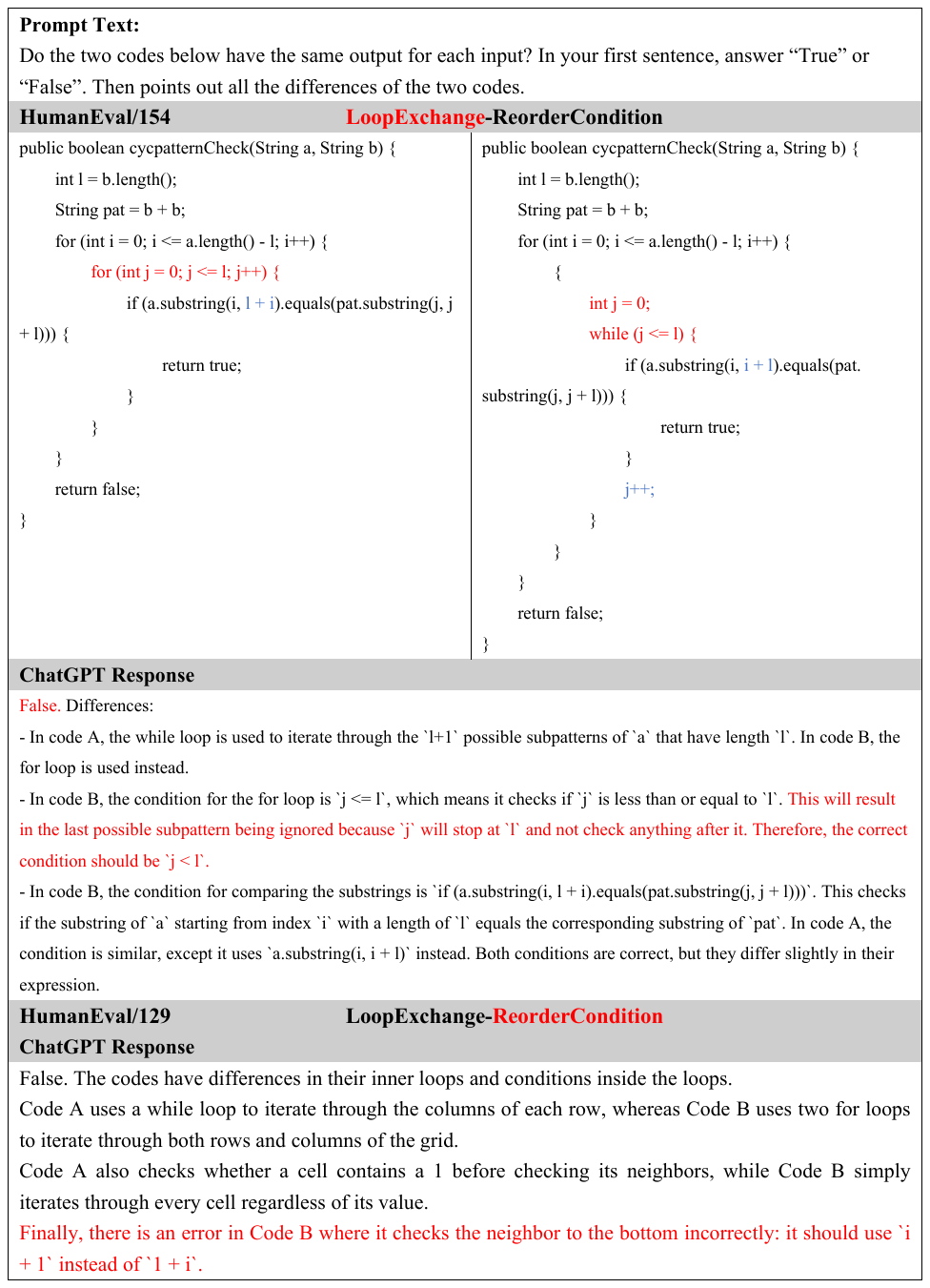}
    \caption{Two failed cases of ChatGPT in the morphism identification experiment. For ``HumanEval/154'', ChatGPT fails to interpret the loop execution condition correctly. For ``HumanEval/129'' ChatGPT reckon equivalent math expression as different. To improve the model's reasoning ability, a possible way is to utilize the method of Chain of Thoughts(CoT) to construct better prompts and give the model some time to think before making a decision. }
    \label{fig:Case_Study_Morphism1}
\end{figure}

\begin{figure}[t]
    \centering
    \includegraphics[width=\linewidth]{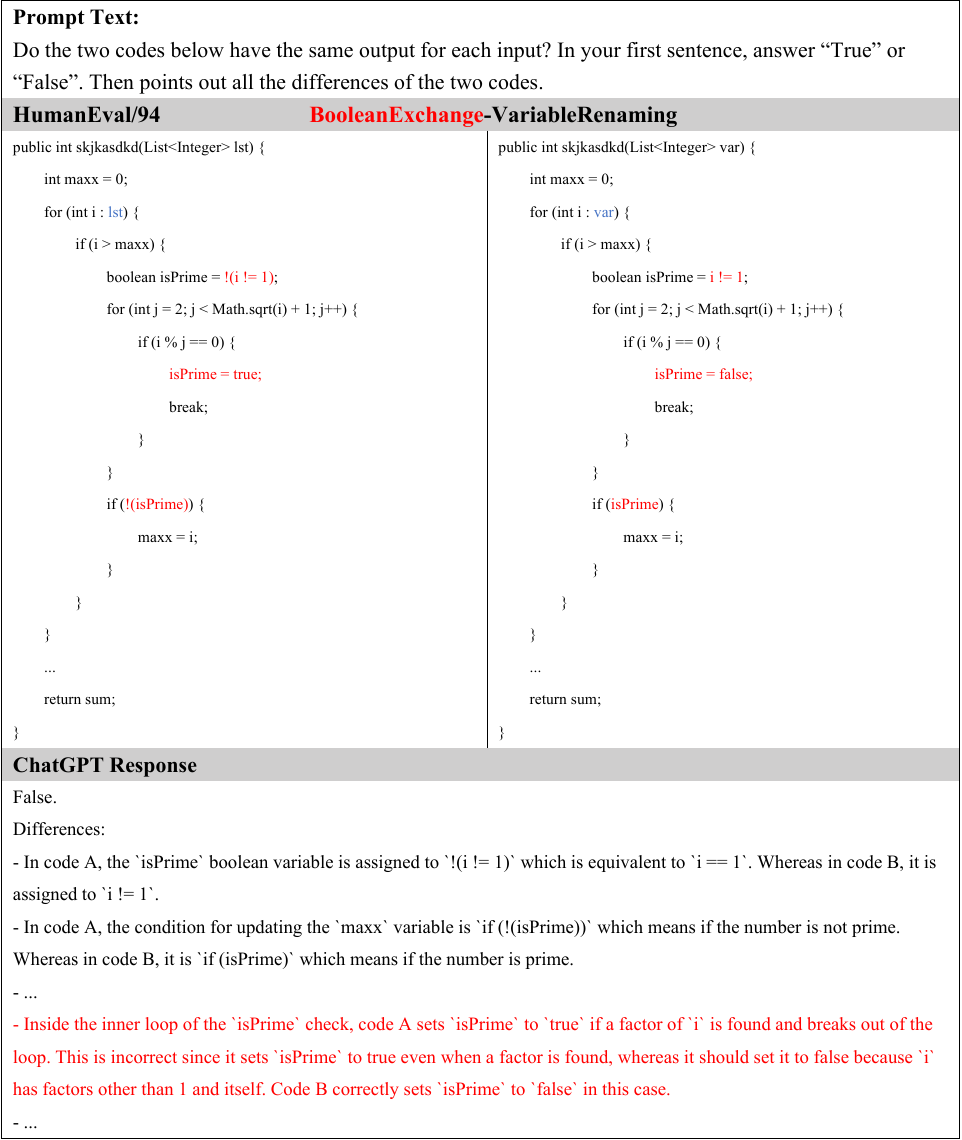}
    \caption{A failed case of ChatGPT in the morphism identification experiment. ChatGPT understands the meaning of the word ``isPrime'' and points out the logical error. Although logical error exists in natural language, the code function is still the same. ChatGPT fails to find that code A is still equivalent to B.}
    \label{fig:Case_Study_Morphism2}
\end{figure}

Refer to Figure~\ref{fig:Case_Study_Morphism1} and Figure~\ref{fig:Case_Study_Morphism2}. Those cases show the model can identify the literate differences between the code pairs but fails to clearly interpret their functional difference.

\subsection{Translation Functor}
\begin{figure}[t]
    \centering
    \includegraphics[width=\linewidth]{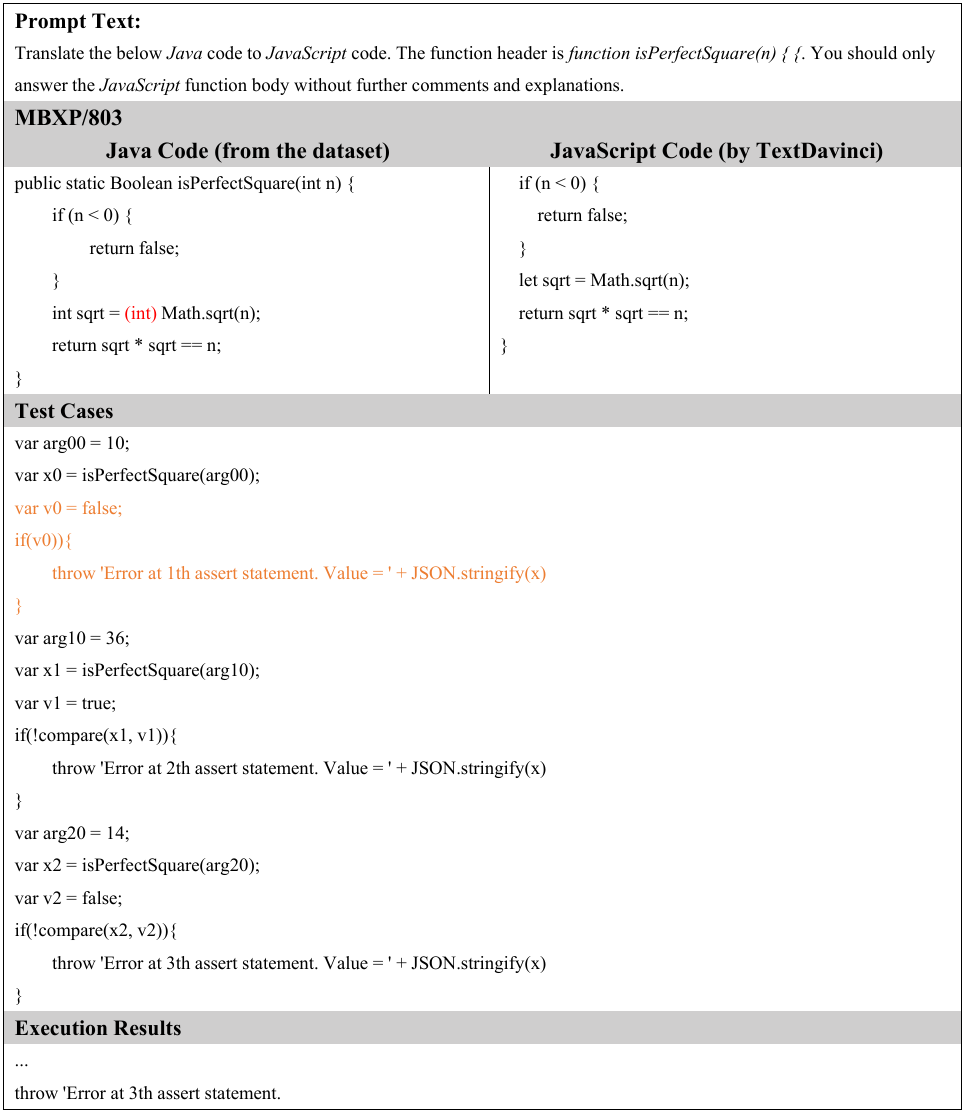}
    \caption{A notable failure case occurred during the translation functor experiment with Text-Davinci. Text-Davinci fails to consider data type conversions during translation. This issue becomes evident when examining the three test cases provided in \texttt{Mxeval}, as the execution encounters an error in the third case. However, if it were not for the incorrect test scripts generated by the model in \texttt{Mxeval}, the translated code should have been identified as incorrect in the first case. This particular case highlights the importance of enhancing the correctness of model-generated execution test scripts for a more accurate test of code.}
    \label{fig:Case_Study_Translation}
\end{figure}

According to the previous statistics, most translation errors happen in the compilation stage. We are also concerned about if there are no compile errors, when will the model make mistakes. Figure~\ref{fig:Case_Study_Translation} shows such a case where some information about the numerical data type is missing when translation. 

\subsection{Explanation Functor and Reproduction Functor}
\begin{figure}
    \centering
    \includegraphics[width=\linewidth]{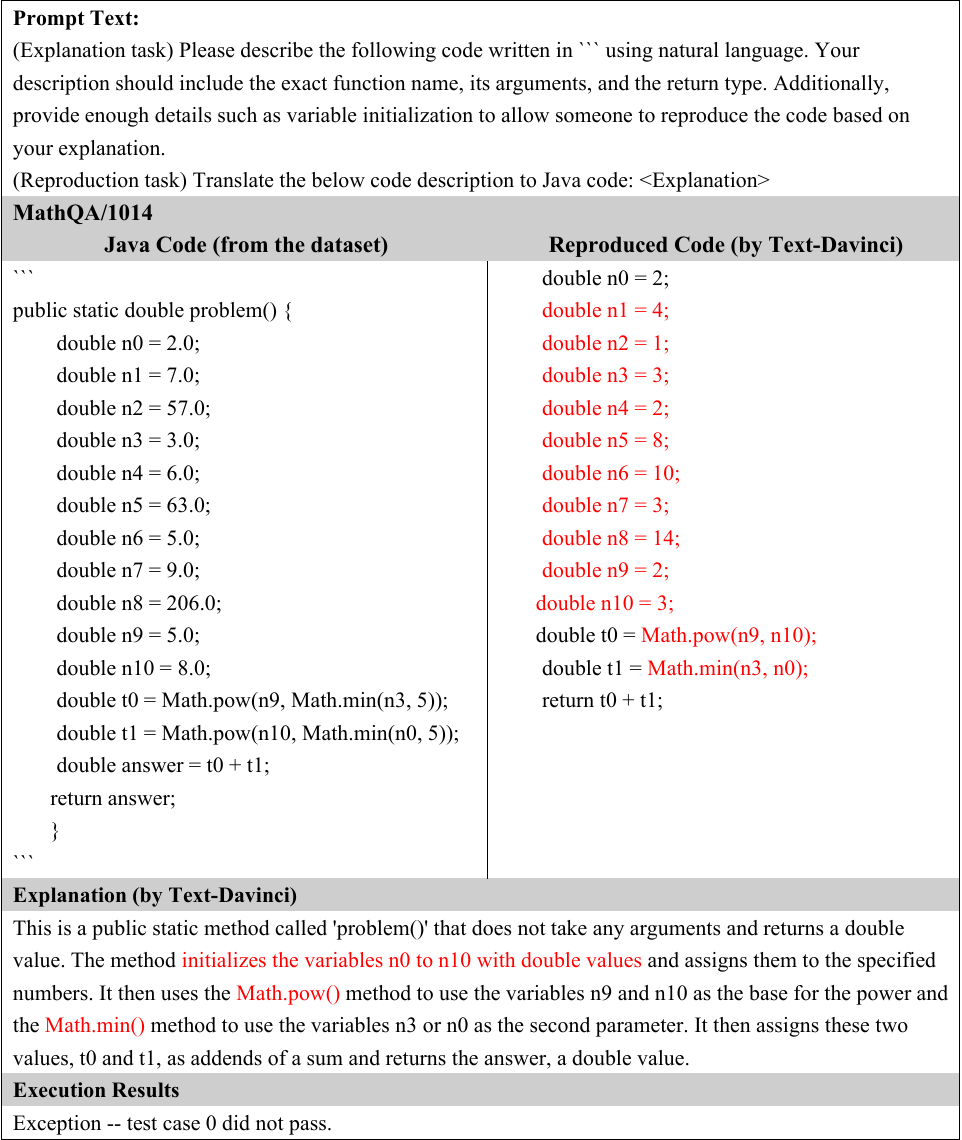}
    \caption{A failure happen in the explanation and reproduction functor experiment. Text-Davinci doesn't explain the code with any numerical details, and it makes up some numbers during reproduction.}
    \label{fig:Case_Study_Reproduction}
\end{figure}

Figure~\ref{fig:Case_Study_Reproduction} shows a case where information loss and hallucination happen during explanation and reproduction. This is often the case when the model deals with the MathQA dataset. To improve the performance of the model, using few-shot prompts to illustrate how to explain the code may be useful.

\end{document}